\documentclass{article}
 \usepackage[preprint]{log_2022}

\usepackage{booktabs}           
\usepackage{multirow}           
\usepackage{amsfonts}           
\usepackage{graphicx}           
\usepackage{duckuments}         

\usepackage[numbers,compress,sort]{natbib}

\usepackage{amsmath,amsfonts,bm}

\def\eqref#1{equation~\ref{#1}}

\def\1{\bm{1}}

\def\mB{{\bm{B}}}

\def\mL{{\bm{L}}}

\DeclareMathAlphabet{\mathsfit}{\encodingdefault}{\sfdefault}{m}{sl}
\SetMathAlphabet{\mathsfit}{bold}{\encodingdefault}{\sfdefault}{bx}{n}

\usepackage{url}
\usepackage[utf8]{inputenc} 
\usepackage[T1]{fontenc}    
\usepackage{url}            
\usepackage{booktabs}       
\usepackage{amsfonts}       
\usepackage{nicefrac}       
\usepackage{microtype}      
\usepackage{xcolor}         
\usepackage{amsmath}
\usepackage{natbib}
\usepackage{adjustbox}
\usepackage{amsmath,amssymb}
\usepackage{graphicx}
\usepackage[utf8]{inputenc}
\usepackage{amsfonts}
\usepackage{graphicx}
\usepackage{textcomp}

\usepackage{algorithm}
\usepackage[noend]{algpseudocode}
\newcommand{\algalign}[2]

\usepackage{cellspace}
\usepackage{graphicx}
\usepackage{caption}
\usepackage{subcaption}

\def\mB{\mbox{$\mathbf{B}$}}

\newenvironment{proof}[1][Proof]{\noindent \textbf{#1.} }{\qedsymbol}
\newcommand{\qedsymbol}{\hspace{\fill}\rule{1.5ex}{1.5ex}}

\newcommand{\beq}{\begin{equation}}
\newcommand{\eeq}{\end{equation}}

\def\mL{\mbox{$\mathbf{L}$}}

\title[Cell Attention Networks]{Cell Attention Networks}

\author[L. Giusti et al.]{%
  Lorenzo Giusti\\
  Sapienza University or Rome\\
  \texttt{lorenzo.giusti@uniroma1.it} \\  
  \And
  Claudio Battiloro\\
  Sapienza University or Rome\\
  \texttt{claudio.battiloro@uniroma1.it} \\
  \And
  Lucia Testa \\
  Sapienza University or Rome\\
  \texttt{lucia.testa@uniroma1.it} \\
  \And
  Paolo Di Lorenzo \\
  Sapienza University or Rome\\
  \texttt{paolo.dilorenzo@uniroma1.it} \\
  \And
  Stefania Sardellitti \\
  Sapienza University or Rome\\
  \texttt{stefania.sardellitti@uniroma1.it}\\
  \And
  Sergio Barbarossa \\
  Sapienza University or Rome\\
  \texttt{sergio.barbarossa@uniroma1.it} 
}

\begin{document}

\maketitle

\begin{abstract}
Since their introduction, graph attention networks achieved outstanding results in graph representation learning tasks. However, these networks consider only pairwise relationships among nodes and then they are not able to fully exploit higher-order interactions present in many real world data-sets. In this paper, we introduce Cell Attention Networks (CANs), a neural architecture operating on data defined over the vertices of a graph, representing the graph as the 1-skeleton of a cell complex introduced to capture higher order interactions. In particular, we exploit the lower and upper neighborhoods, as encoded in the cell complex, to design two independent masked self-attention mechanisms, thus generalizing the conventional graph attention strategy. The approach used in CANs is hierarchical and it incorporates the following steps: i) a lifting algorithm that learns {\it edge features} from {\it node features}; ii) a cell attention mechanism to find the optimal combination of edge features over both  lower and upper neighbors; iii)  a hierarchical {\it edge pooling} mechanism to extract a compact meaningful set of features. The experimental results show that CAN is a low complexity strategy that compares favorably with state of the art results on graph-based learning tasks.
\end{abstract}

\section{Introduction}

Graph Neural Networks (GNNs) find applications in a plethora of fields, like computational chemistry~\cite{gilmer2017neural}, social networks~\cite{fan2019graph} and physics simulations~\cite{shlomi2020graph}. Since their introduction~\cite{gori2005new, scarselli2008graph}, GNNs have shown remarkable results in learning tasks when data are defined over a graph domain, where the flexibility of neural networks is coupled with prior knowledge about data relationships, expressed in terms of the underlying topology. The literature on GNNs is large and numerous techniques have been studied, usually categorized in spectral~\cite{Bruna19, kipf2016semi} and non-spectral methods~\cite{hamilton2017inductive, DuvenaudMABHAA15, atwood2016diffusion}. 

Generally speaking, the idea is learning representations of node attributes using local aggregation, where the neighborhood is formally represented by the graph topology. By leveraging this basic but powerful idea, outstanding performance has been achieved in many traditional tasks such as classification for nodes or entire graphs~\cite{kipf2016semi, velivckovic2017graph, hamilton2017inductive} or link prediction~\cite{zhang2018link} as well as more specialized ones such as \emph{protein folding}~\cite{jumper2021highly} and \emph{neural algorithmic reasoning}~\cite{davies2021advancing, velivckovic2021neural}. At the same time, a major performance boost to deep learning algorithms has been offered by the inclusion of attention mechanisms, introduced to handle sequence-based tasks~\cite{bahdanau2014neural}, enabling for variable sized inputs, to concentrate on their most important features. Then, pioneering works introduced  graph attention networks~\cite{velivckovic2017graph, yun2019graph, EdgeRibeiro} achieving state-of-the-art results in most of the aforementioned tasks. 

Graphs can be also seen as a simple instance of a \emph{topological space}, able to capture {\it pairwise} interactions through the presence of an edge between any pair of directly interacting nodes. However, despite their overwhelming popularity, graph-based representations are unable to consistently model higher order relations, which play a crucial role in many practical applications. Examples where multiway relations cannot be reduced to an ensemble of pairwise relations are gene regulatory networks~\cite{sever2015signal,lambiotte2019networks}, where some reactions occur only when a set of multiple (not only two) genes interact, or applications in network neuroscience~\cite{giusti2016two, sizemore2018cliques}.
To overcome this problem, many architectures defined on general hypergraphs have been proposed~\cite{chien2022you,Yadati2019, hypergraphneuarlnet, Zhang2020Hyper-SAGNN:,NEURIPS2020_217eedd1}, and some works incorporating attention mechanisms for hypergraphs neural networks have been published~\cite{HyperAttMultimodal, HyperAttNetworks}.

However, in the meanwhile, pioneering works on Topological Signal Processing~\cite{barbarossa2020topological, sardellitti2022cell} demonstrated the benefit of processing signals defined on simplicial and cell complexes, which are specific examples of hyper-graphs with a rich algebraic description, and can easily encode multi-way relationships hidden in the data in a low-complexity fashion. Consequently, there was a natural interest in the design of (deep) neural networks architectures able to learn from data defined on simplicial and cell complexes, as summarized in the following state of the art.\\

\subsection{Related Works}
Despite Topological Deep Learning is an emerging research area that has been introduced quite recently, numerous pioneering works appeared in this field. In~\cite{bodnar2021weisfeiler}, message passing neural networks (MPNNs)~\cite{gilmer2017neural} are adapted to simplicial complexes (SCs), and a Simplicial Weisfeiler-Lehman (SWL) colouring procedure for distinguishing non-isomorphic SCs is introduced. The aggregation and updating functions are able to process the data exploiting lower and upper neighbourhoods. The architectures in~\cite{bodnar2021weisfeiler}, namely Message Passing Simplicial Networks (MPSNs), are a generalization of Graph Isomorphism Network (GIN)~\cite{xu2018powerful}.  In~\cite{bodnarcwnet}, message passing neural networks able to handle data defined over regular cell complexes are introduced under the name of CW Networks (CWNs), and they are proven to be not less powerful than the 3-WL test. The work in (~\cite{sheaf2022}) introduced Neural Sheaf Diffusion Models, neural architectures grounded in the theory of cellular sheaves able to improve learning performance on graph-based tasks especially in heterophilic settings. In~\cite{giusti22}, a novel attention neural architecture that operates on data defined on simplicial complexes leveraging masked self-attention layers is introduced, taking into account lower and upper neighbourhoods and introducing a proper technique to process the harmonic component of the data based on the design of a sparse projection operator. A similiar architecture is proposed in~\cite{anonymous2022SAT}, which re-weights the interactions between neighbouring simplices through an orientation equivariant attention mechanism. However, none of these works considered masked self-attention mechanisms for architectures designed to handle data defined over cell complexes. Finally, the work in ~\cite{hajij2022} introduced a broad class of attentional architectures operating on generalized higher-order domains called Combinatorial Complexes. 

\subsection{Contribution}

The aim of this paper is to introduce \textit{cell attention networks}, i.e., a fully  attentional low-complexity architecture able to learn from data defined over the nodes of a graph by incorporating the graph within a cell complex and working at the edge-level in order to extract higher order interactions. The idea is to implement an attention mechanism that handles relations among nearby edges, where the neighborhood is formally represented by a cell complex. To this aim, we exploit a hierarchical approach that lifts up {\it node features} to derive {\it edge features}, and then it computes the attention coefficients between nearby edges to find the optimal combination of edge features. In particular, we devise an \emph{attentional lift mechanism} that learns the data over the edges of the complex leveraging a self-attention mechanism over the features of the vertices that are on their boundaries. Consequently, our architecture is also equipped with a cellular lifting map algorithm that embeds the graph domain into a regular cell complex (CW).   Other graph lifting techniques have been used in previous works, such as  clique complexes ~\cite{Ferri2018,Milo2002}, or more sophisticated structures based on incidence tensors ~\cite{albooyeh-etal-2020-sample}.Finally, please notice that the proposed  architecture presents nice features of explainability due to its fully attention-driven design: by simply inspecting the attention coefficients, it is possible to understand the contribution of the single cells for the learning task. For instance, in a computational chemistry task, the upper attention coefficients may represent the importance of the interaction between two atoms inside the rings of a molecule. Also, considering a network traffic problem, the pooling attention coefficients may tell us which links can be considered obsolete, or even detect communities in social networks. 

\section{Cell Complexes}

In this section we recall the basics of regular cell complexes,  which are topological spaces provided with a rich algebraic structure that enables an efficient representation of high-order interaction systems, and we explain their relation with usual graphs $\mathcal{G}=(\mathcal{V},\mathcal{E)}$. 
In particular, we will first introduce the  definition of a regular cell complex, then we will describe few additional properties enabling the representation of cell complexes via boundary operators.

\textit{\textbf{Definition 1 (Regular cell complex) }~\cite{hansen2019toward, bodnarcwnet}. A {\it regular cell complex} is a topological space $\mathcal{C}$ together with a partition $\{\mathcal{X}_{\sigma}\}_{\sigma \in \mathcal{P}_{\mathcal{C}}}$ of subspaces $\mathcal{X}_{\sigma}$ of $\mathcal{C}$ called $\mathbf{cells}$, where $\mathcal{P}_{\mathcal{C}}$ is the indexing set of $\mathcal{C}$, such that}

\begin{enumerate}
    \item For each $c$ $\in$  $\mathcal{C}$, every sufficient small neighborhood of $c$ intersects finitely many $\mathcal{X}_{\sigma}$;  
    \item For all $\mathcal{\tau}$,$\mathcal{\sigma}$ we have that $\mathcal{X}_{\tau}$ $\cap$ $\overline{\mathcal{X}}_{\sigma}$ $\neq$ $\varnothing$ iff $\mathcal{X}_{\tau}$ $\subseteq$ $\overline{\mathcal{X}}_{\sigma}$, where $\overline{\mathcal{X}}_{\sigma}$ is the closure of the cell;
    \item Every $\mathcal{X}_{\sigma}$ is homeomorphic to $\mathbb{R}^{n_{\sigma}}$ for some $n_{\sigma}$;
    \item For every $\sigma$ $\in$ $\mathcal{P}_{\mathcal{C}}$ there is a homeomorphism $\phi$ of a closed ball in $\mathbb{R}^{{n}_{\sigma}}$ to $\overline{\mathcal{X}}_{\sigma}$ such that the restriction of $\phi$ to the interior of the ball is a homeomorphism onto $\mathcal{X}_{\sigma}$.
\end{enumerate}

Condition (2) implies that the indexing set $P_{\mathcal{C}}$ has a poset structure, given by $\tau$ $\leq$ $\sigma$ iff $\mathcal{X}_{\tau}$ $\subseteq$ $\overline{\mathcal{X}_\sigma}$. This is known as the face poset of $\mathcal{C}$. The regularity condition (4) implies that all topological information about $\mathcal{C}$ is encoded in the poset structure of $P_{\mathcal{C}}$. Then, a regular cell complex can be identified with its face poset.

\textit{\textbf{Definition 2 (k-skeleton)}. A {\it k-skeleton} of a cell complex $\mathcal{C}$, denoted $\mathcal{C}^{(k)}$, is the subcomplex of $\mathcal{C}$ consisting of cells of dimension at most k.}

From Definition 1 and Definition 2, it is trivial to check the 0-skeleton of a cell complex is the set of vertices $\mathcal{C}^{(0)} = \mathcal{V}$ and the 1-skeleton is the underlying graph $\mathcal{C}^{(1)}=\mathcal{G}(\mathcal{V},\mathcal{E)}$; we refer to 2-cells as polygons and, in general, there is little interest with dimensions above two. Regular cell complexes can be described via an incidence relation (boundary relation) with a  reflexive and transitive closure that is consistent with the partial order introduced in Definition 1. The boundary relation describes which cells are on the boundary of other cells.

\textit{\textbf{Definition 3 (Boundary relation)}.We have the boundary relation  $\sigma$ $\prec$ $\tau$ iff $\dim({\sigma})$ $\leq$ $\dim({\tau})$ and there is no cell $\delta$ such that
$\sigma$ $\leq$ $\delta$ $\leq$ $\tau$ .}

We can use the previous definitions to define the four types of (local) adjacencies present in cell complexes, following the approach from~\cite{bodnarcwnet}: 

\textit{\textbf{Definition 4 (Cell complex adjacencies) }~\cite{bodnarcwnet}. For a cell complex $\mathcal{C}$ and a cell $\sigma \in \mathcal{P}_{\mathcal{C}}$, we define: }

\begin{itemize}
    \item The boundary adjacent cells $\mathcal{B}(\sigma)$ $=$ $\{ \tau $ $|$ $\tau \prec \sigma \}$, are the lower-dimensional cells on the
boundary of $\sigma$. For instance, the boundary cells of an edge are its vertices and the boundary cells of a polygon are its edges.

    \item The co-boundary adjacent cell $\mathcal{CB}(\sigma)$ $=$ $\{\tau $  $|$ $  \sigma \prec \tau \}$, are the higher-dimensional cells with
    $\sigma$ on their boundary. For instance, the co-boundary cells of a vertex are the edges having that vertex as an endpoint and the co-boundary of an edge are the polygons having that edge as one of its sides. 
    
    \item The lower adjacent cells $\mathcal{N}_{d}(\sigma)$ $=$ $\{ \tau $ $|$ $ \exists \delta$ such that $\delta \prec \sigma$ and $\delta \prec \tau\}$, are the cells of
the same dimension as $\sigma$ that share a lower dimensional cell on their boundary. The line graph
adjacencies between the edges are a classic example of this.

     \item The upper adjacent cells $\mathcal{N}_{u}(\sigma)$ $=$ $\{ \tau $ $|$ $ \exists \delta$ such that $\sigma \prec \delta$ and $\tau \prec \delta\}$. These are the cells of
the same dimension as $\sigma$ that are on the boundary of the same higher-dimensional cell as $\sigma$. 
\end{itemize}

\begin{figure}[t]
    \centering
    \includegraphics[width=.42\textwidth]{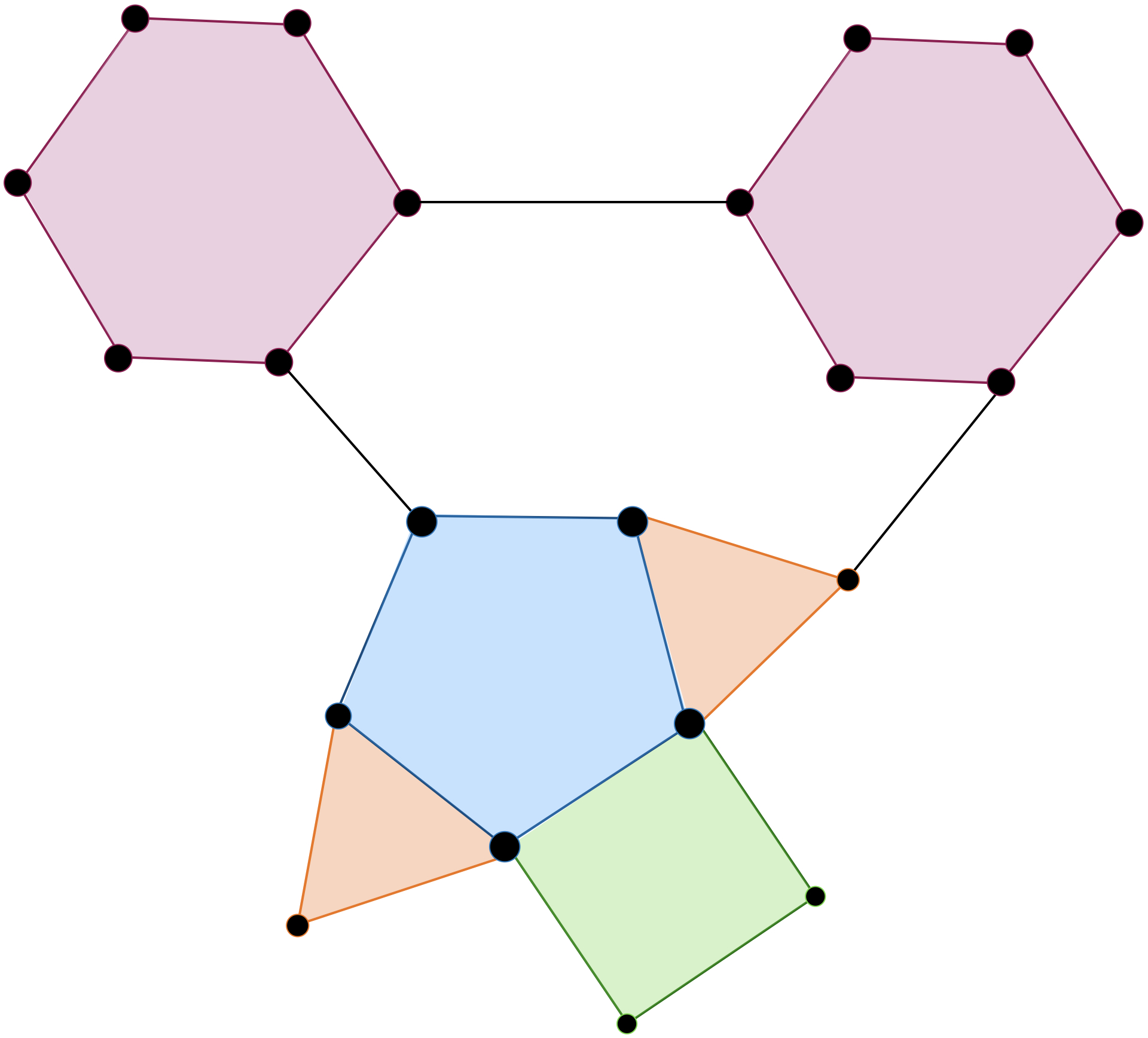}
    \caption{Illustration of a geometric cell complex $\mathcal{C} = (\mathcal{V}, \mathcal{E}, \mathcal{P})$. In this figure, it is possible to distinguish the topology of $\mathcal{P}$ by the color attached to its elements,i.e., the polygons. Two polygons share the same color if the have the same number of boundary elements, i.e., triangles are orange, squares are green, and so on.}
    \label{fig:cc}
\end{figure}

Finally, it is possible to incorporate a graph $\mathcal{G}$ in a higher order cell complex $\mathcal{C}_\mathcal{G}$ by attaching polygons to closed paths of edges having no internal chords. 

\subsection{Algebraic Representation}

Let us now introduce an algebraic representation of the cell complexes based on the incidence relations among cells. We need first to introduce
the orientation of a cell complex
by generalizing the concept of orientation of a simplex~\cite{bodnar2021weisfeiler, grady2010discrete}. To define the orientation of a $k$-cell, we may apply a simplicial decomposition~\cite{grady2010discrete}, which consists in  subdividing the cell into a set of internal $k$-simplices, so that, by orienting a single internal simplex, the orientation propagates to the entire cell. Defining the transposition as the permutation of two elements, two orientations are equivalent if each of them can be recovered from the other through an even number of transpositions. 
Given an oriented cell complex ${\cal C}$, let ${S}_{k}$ denote the number of cells in dimension $k$, and $\tau_i$ and $\sigma_j$ denote two cells of the complex with dimension $\dim(\sigma_j) = k$ and $\dim(\tau_i) = k-1$ respectively. 
The $k-th$ signed boundary matrix $\mathbf{B}_{k} \in \mathbb{R}^{S_{k-1} \times S_{k}}$ of ${\mathcal{C}}$ is:

  \beq \label{inc_coeff}
  [\mathbf{B}_k]_{i,j}=\left\{\begin{array}{rll}
  1, & \text{if} \; \tau_i \prec_{+} \sigma_j \\
  -1,& \text{if} \; \tau_i \prec_{-} \sigma_j\\
  0,&  \; \text{otherwise}\\
  \end{array}\right. 
  \eeq
 
where we use the notation $\prec_{+}$ to indicate coherent orientation between two cells and $\prec_{-}$
to indicate a opposite orientation between two cells. 

As a particular case, let us consider a cell complex of order two  $\mathcal{C}=\{\mathcal{V},\mathcal{E},\mathcal{P}\}$, where $\mathcal{V}$, $\mathcal{E}$, $\mathcal{P}$ denote the set of  $0$, $1$ and $2$-cells, i.e., vertices, edges and polygons, respectively. We denote their cardinality by $|\mathcal{V}|=V$, $|\mathcal{E}|=E$ and $|\mathcal{P}|=P$. Then, the two incidence matrices describing the connectivity of the complex are $\mB_1 \in \mathbb{R}^{V\times E}$ and $\mB_2 \in \mathbb{R}^{E\times P}$, where $\mB_2$ can be written as in~\cite{sardellitti2022cell}: $\mB_2=[\mB_{T},\mB_{Q},\ldots,\mB_{P_{S}} ]$, with  $\mB_{T}$, $\mB_{Q}$ and  $\mB_{P_{S}}$ denoting the incidences between edges and, respectively, triangles, quadrilaterals, up to polygons with $P_{S}$ sides (where each polygon does not include any internal chord between any pair of its vertices).  
 An interesting property of the incidence matrices is that $\mB_k \mB_{k+1}=\mathbf{0}$, for all $k$. 

Finally, the structure of a $K$-cell complex can be described through the {\it higher-order combinatorial Laplacians} defined as~\cite{eckmann1944harmonische, Lim}: 
\begin{align}
    &\mL_0=\mB_1\mB_1^T, \nonumber\\
    &\mL_k=  \mL_{k}^{d}+\mL_{k}^{u} ,\;\;\qquad \hbox{$\;\;$ for $k=1,\ldots,K-1$,} \nonumber\\
    &\mL_K=\mB_K^T\mB_K,
\end{align}
where $\mL_{k}^{d} =\mB_{k}^T \mB_{k}$ and  $\mL_{k}^{u} =\mB_{k+1} \mB_{k+1}^T$ are respectively the lower and upper Laplacians, encoding the lower and upper adjacencies  of the $k$-order cells.
Note that  $\mathbf{L}_{0}$  corresponds to the combinatorial Laplacian used for graph representations.

\subsection{Data over Cell Complexes}

Let $\mathcal{C}_k$ be the set of k-th order cells in a cell complex $\mathcal{C}$. In most of the cases the focus is on complex $\mathcal{C}^{(2)}$ of order up to two, thus a set of vertices $\mathcal{V}$ with $|\mathcal{V}| = V$, a set of edges $\mathcal{E}$ with $|\mathcal{E}|=E$ and a set of polygons $\mathcal{P}$ with $|\mathcal{P}| = P$ are considered, resulting in ${\cal C}_{0}={\cal V}$ (cells of order 0), ${\cal C}_{1}={\cal E}$ (cells of order 1) and ${\cal C}_{2}={\cal P}$ (cells of order 2). In Fig. \ref{fig:cc} we sketch an example of a cell complex of order $2$.

A $k$-cell signal is defined as a mapping from the set of all $k$-cells contained in the complex to real numbers:
\begin{equation} \label{gen_sig}
\mathbf{x}_{k}: {\cal C}_{k} \rightarrow \mathbb{R}, \,\,\quad k=0, 1, \ldots K.
\end{equation}
The order of the signal is one less the cardinality of the elements of ${\cal C}_{k}$. Therefore, for a complex $\mathcal{C}^{(2)}$, the $k$-cell signals are defined as the following mappings: 
\begin{equation}
\mathbf{x}_{0}: {\cal V} \rightarrow \mathbb{R} , \qquad \mathbf{x}_{1}: {\cal E} \rightarrow \mathbb{R} , \qquad \mathbf{x}_{2}: {\cal P} \rightarrow \mathbb{R} ,
\end{equation}
representing vertex, edge and polygon signals, respectively. 

In this work, we will consider only vertex and edge signals. In particular, we will refer to an instance of the former as $\mathbf{x}_{i}$ while the instance of the latter will be referred as $\mathbf{x}_{e}$.

\section{Cell Attention Networks}\label{sec:can}

The aim of this work is to extend Graph Attention Networks introduced in~\cite{velivckovic2017graph} to account for multi-way relationships, i.e., performing a masked self-attention mechanisms at the edge level. The proposed hierarchical architecture, which we refer to as Cell Attention Network (CAN) (\ref{fig:tan}), starts with the embedding of the input graphs in regular cell complexes via a skeleton-preserving cellular lifting map and an attentional lift procedure enabling the derivation of edge features from node features. Then, we introduce a novel edge-level attentional message passing scheme. After each round of message passing, we perform a novel edge pooling operation and a local readout to reduce complexity; finally, after the last message-passing round, a global readout is applied. As in the previous sections, we denote the input graph(s) with $\mathcal{G}=(\mathcal{V},\mathcal{E)}$ and the input node features of node $i \in \mathcal{V}$ with $\mathbf{x}_i \in \mathbb{R}^{F_n}$.

\subsection{Cellular Lifting Map}

We first need to incorporate input graphs in  regular cell complexes. To address this challenge, we exploit the notion of skeleton-preserving cellular lifting map presented in~\cite{bodnarcwnet} and defined as:

\textit{\textbf{Definition 5 (Skeleton-Preserving Cellular Lifting Map) }~\cite{bodnarcwnet}. A cellular lifting map $s: \mathcal{G} \rightarrow \mathcal{C}_{\mathcal{G}}$ is a skeleton preserving function that incorporates a graph $\mathcal{G}$ into a regular cell complex $\mathcal{C}_{\mathcal{G}}$, such that, for any graph $\mathcal{G}$, the 1-skeleton (i.e., the underlying graph) of  $s(\mathcal{G})$ and $\mathcal{G}$ are isomorphic .}

Informally, Definition 5 just requires that the lifting map keeps the underlying graph structure unchanged. Several cellular lifting map can be exploited, in this work we opted for a lifting map that attach cells to all the induced (or chordless) cycles, where $k$ can be considered a hyperparameter to be chosen arbitrarily, and which controls the maximum size of the polygons (2-cells) of the complex.


\subsection{Attentional lift}
 After the Cellular Lifting Map, we need to learn edge features, thus performing a lift operation on the node features that we refer to as \emph{attentional lift}. To this aim, we exploit a masked multi-head self-attention mechanism~\cite{velivckovic2017graph}. The procedure is based on the computation of $F^0$ attention heads such that, for each pair of nodes $i$,$j \in \mathcal{V}$ connected by an edge $e \in \mathcal{E}$, the corresponding edge features $\mathbf{x}_{e} \in \mathbb{R}^{F^0}$ are given by the concatenation of the resulting attention scores.

\textit{\textbf{Definition 6 (Attentional Lift)}}. An Attentional Lift is a \emph{learnable} function $g: \mathbb{R}^{F_n} \times \mathbb{R}^{F_n} \rightarrow \mathbb{R}^{F^0}$,  of the form: 
\begin{equation}\label{eq:attentional_lift}
    \mathbf{x}_{e} =g(\mathbf{x}_i,\mathbf{x}_j)= \overset{F^0}{\underset{k=1}{||}}a_n^k(\mathbf{x}_i,\mathbf{x}_j), \quad \forall e \in \mathcal{E}.
\end{equation}
where $a_n^k: \mathbb{R}^{F_n} \times \mathbb{R}^{F_n}  \rightarrow \mathbb{R}$ is the $k$-th (shared across nodes) learnable attention function, and $||$ is the concatenation operator. Since the order of the nodes connected by an edge should not change the corresponding lifted edge features, we assume that the functions $a_n^k$ are symmetric. 

\subsection{Cell Attention}

\begin{figure}[t]
    \centering
    \includegraphics[width =.8\textwidth]{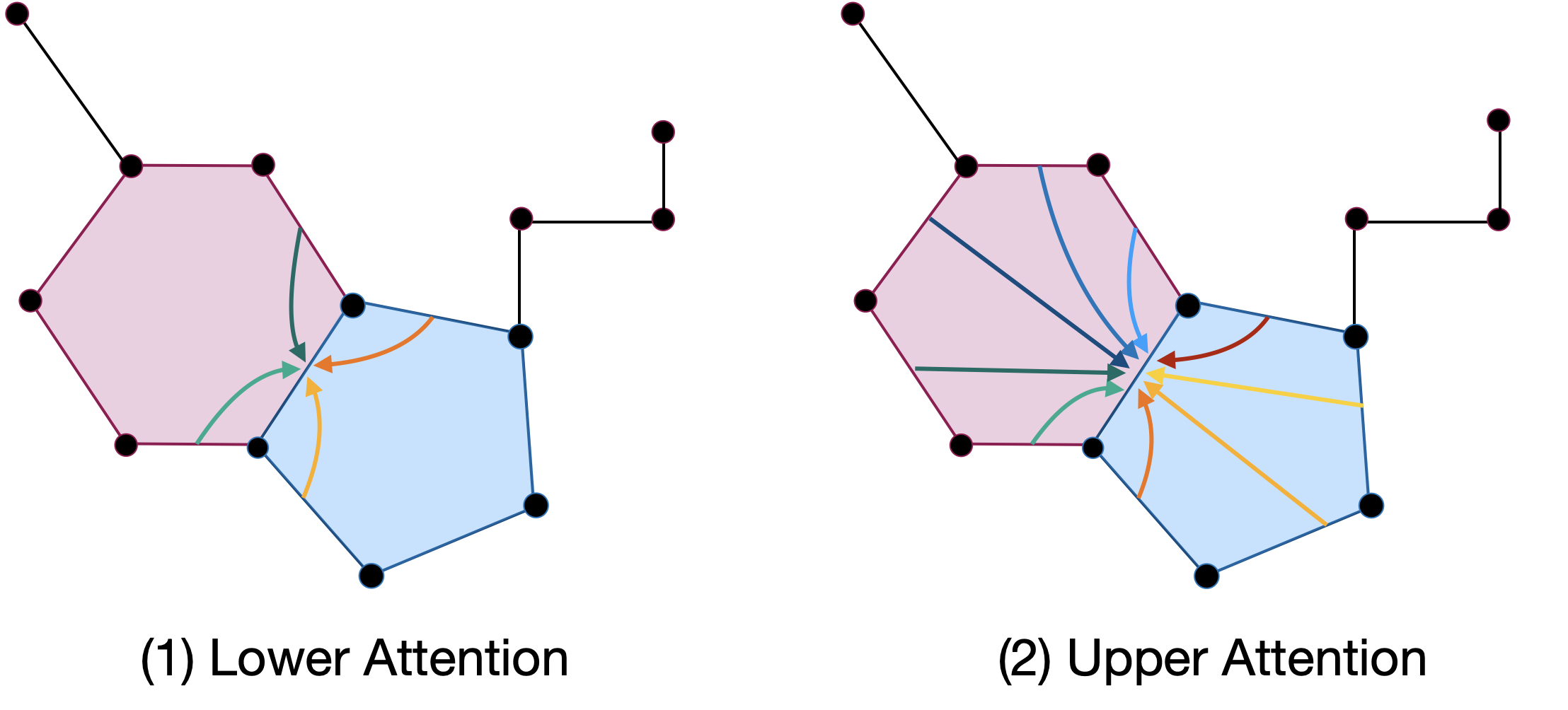}
    \caption{Illustration of upper and lower attention}
    \label{fig:att}
\end{figure}

In this section we introduce Cellular Attentional Message-passing, an attentional message passing scheme operating at edges level on the learned edges features of Eq. (\ref{eq:attentional_lift}) exploiting the connectivity given by the regular cell complex $\mathcal{C}_{\mathcal{G}}$ computed via the cellular lifting map of Definition 5. Before describing the proposed scheme, please notice that, as previously introduced, we will perform an edge pooling operation after the message-passing round at each layer $l \in \{1,...,L\}$, meaning that the architecture will produce a sequence of cell complexes $\{\mathcal{C}^l\}_l$ such that  $\mathcal{C}^{l+1} \subseteq \mathcal{C}^{l}$ (due to the fact that the corresponding edge sets are such that $\mathcal{E}^{l+1} \subseteq \mathcal{E}^{l}$); we will describe the edge pooling in details in the next section. As already introduced in Section 2, there are two types of adjacencies that can be exploited when dealing with cell complexes. In particular, since the message exchange happen at the edges level, in each layer $l$, our message-passing scheme exploits upper and lower edge adjacencies $\mathcal{N}^l_{d}(e)$ and $\mathcal{N}^l_{u}(e)$, associated with the cell complex $\mathcal{C}_{l}$, $l=1,\ldots,L.$ At each layer $l$, we introduce a learnable upper attention function $a^l_{u}: \mathbb{R}^{F^l}  \times \mathbb{R}^{F^l}  \rightarrow \mathbb{R}$, responsible to evaluate the reciprocal importance of two edges that are part of the same polygon, and a lower attention function $a^l_{d}: \mathbb{R}^{F^l} \times \mathbb{R}^{F^l}  \rightarrow \mathbb{R}$, responsible to evaluate the reciprocal importance of two edges that share a common node. Therefore, edges embedding are updated in the $l-th$ message passing round as:

\begin{align}\label{message_passing_scheme}
    \widetilde{\mathbf{h}}_{e}^{l} = \phi^l\Bigg(\mathbf{h}_{e}^{l}, \, \bigoplus_{k \in \mathcal{N}^l_d(e)} a^l_d(\mathbf{h}_{e}^l,\mathbf{h}_{k}^l)\, \psi_d^l(\mathbf{h}_{k}^l), \, \bigoplus_{k \in \mathcal{N}^l_u(e)} a^l_u(\mathbf{h}_{e}^l,\mathbf{h}_{k}^l)\, \psi_u^l(\mathbf{h}_{k}^l) \Bigg) \in \mathbb{R}^{F^{l+1}}, \; \forall e \in \mathcal{E}^l,
\end{align}

where $\bigoplus$ is any permutation invariant (aggregation) operator (e.g., sum, mean, max, ...), $\phi^l$ is a (possibly) learnable function, $\psi_u^l$ and $\psi_d^l$  are  learnable functions sharing the weights with $a^l_u$ and $a^l_d$, respectively (as usual in attentional settings),  $\mathbf{h}_{e}^{0} = \mathbf{x}_{e}$, $\mathcal{C}^0 = \mathcal{C}_{\mathcal{G}}$ (thus $\mathcal{E}^0 = \mathcal{E}$). Obviously, multi-head attention can be trivially injected following the usual concatenation or averaging approach~\cite{giusti22,velivckovic2017graph}. A pictorial example of how upper and lower attention work is depicted in Figure \ref{fig:att}.

\subsection{Edge Pooling} 

\begin{figure}[t]
    \centering
    \includegraphics[width=\textwidth, height = 4cm]{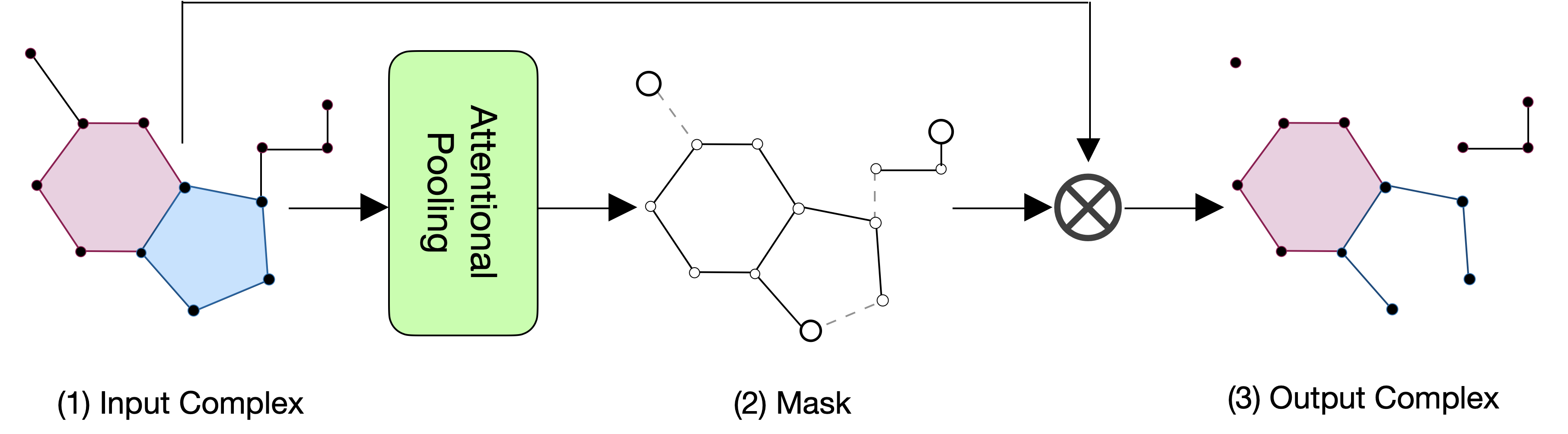}
    \caption{Illustration of the proposed edge pooling procedure}
    \label{fig:pooling}
\end{figure}

 In this section, we present a self-attention edge pooling technique, adopting a variation of the method used in~\cite{lee2019self}. Let $\widetilde{\mathbf{h}}_{e}^{l} \in \mathbb{R}^{F^{l+1}}$ be the hidden feature vector associated to edge $e$ obtained via attentional message-passing after the $l$-th message-passing round. The edge attention pooling operation consists in computing a self-attention score $\gamma_{e}^{l} \in \mathbb{R}$ for each edge of the complex via a pooling learnable attention function $a^l_{p}: \mathbb{R}^{F^{l+1}} \rightarrow \mathbb{R}$ : 
\begin{equation}\label{pooling_scores}
    \gamma_{e}^{l} = a^l_p \left(\widetilde{\mathbf{h}}_{e}^{l} \right) \qquad \forall e \in \mathcal{E}^l.
\end{equation}
Let $k \in (0, 1]$ be the \emph{pooling ratio}, i.e., the fraction of the edges that will be retained over the number of edges in input to the self-attention edge pooling layer. At this point, we keep the $\lceil k |\mathcal{E}^l| \rceil$ edges belonging to the set  $\mathcal{E}^{l+1} = \{ e  : e \in \mathcal{E}^l \textrm{ and } \gamma_e^{l} \in \text{top-k}(\{\gamma_{e}^{l}\}_{e \in \mathcal{E}^l}, \lceil k |\mathcal{E}^l| \rceil)  \} \subseteq \mathcal{E}^l$ where $\text{top-k}(\{\gamma^l_e\}_{e \in \mathcal{E}^l}, \lceil k |\mathcal{E}^l| \rceil)$ is the set of  the highest $\lceil k |\mathcal{E}^l| \rceil$ self-attention scores. Finally, the  feature vectors that will be kept after the pooling stage are scaled as:
\begin{equation}\label{pooling_scaling}
    \mathbf{h}_{e}^{l+1} =  \gamma_{e}^{l} \widetilde{\mathbf{h}}^{l}_{e} , \;\; \forall  e  \in \mathcal{E}^{l+1}.
\end{equation}
After the edge pooling, we consequently need  to adjust the structure of the cell complex $\mathcal{C}^l$ to obtain a consistent updated complex $\mathcal{C}^{l+1}$ . 
To this aim, we apply the procedure depicted in Fig. \ref{fig:pooling}: If an edge $e$ belongs to $\mathcal{E}^{l}$ but  is not contained in $ \mathcal{E}^{l+1}$, the lower connectivity is updated by disconnecting the nodes that are on the boundary of $e$, while the upper connectivity is updated by removing the polygons that have $e$ on their boundaries.

Finally, we considered also a hierarchical version of the aforementioned self-attention edge pooling operation as in~\cite{cangea2018towards}. To this aim we employ a (by-layer) readout operation on the hidden feature  $\{\mathbf{h}_e^{l+1}\}_{e \in \mathcal{E}^{l+1}}$ to obtain an aggregate embedding of the whole complex $\mathcal{C}^{l+1}$ as:
\begin{equation}\label{local_readout}
    \mathbf{h}_{\mathcal{C}^{l+1}} = \bigoplus_{e \in \mathcal{E}^{l+1}}  \mathbf{h}_{e}^{l+1}.
\end{equation} 
Then, after the last hidden layer, a final (global) readout operation is performed, e.g., by aggregating all the previously computed complexes embeddings: 
\begin{equation}\label{global_readout}
\mathbf{h}_{\mathcal{C}} = \bigoplus_{l} \mathbf{h}_{\mathcal{C}^{l}}.
\end{equation}
Finally, the result of the final aggregation is fed to a multi-layer perceptron (MLP) if needed for the learning task.

\subsection{CAN Architecture and Symmetries}
\begin{figure}[t]
    \centering
    \includegraphics[width=\textwidth, height = 4cm]{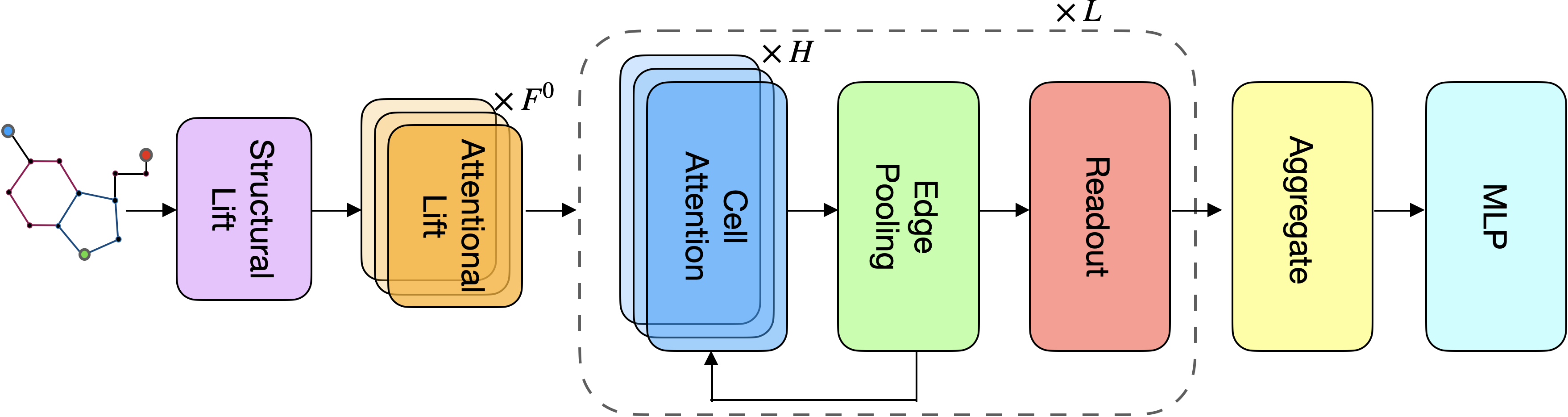}
    \caption{Illustration of a cell attention network}
    \label{fig:tan}
\end{figure}

In summary, a \emph{Cell Attention Network} with input graph(s) $\mathcal{G}=(\mathcal{V},\mathcal{E})$ and input node features $\{\mathbf{x}_i\}_{i \in \mathcal{V}}$ is defined as the stack of: (i) a skeleton-preserving cellular lifting map to obtain a regular cell complex $\mathcal{C}_{\mathcal{G}}$ from a graph $\mathcal{G}$; (ii) a multi-head attentional lift to obtain edge features $\{\mathbf{x}_e\}_{e \in \mathcal{E}}$ from node features $\{\mathbf{x}_i\}_{i \in \mathcal{V}}$ ; (iii) a stack of $L$ cell attention layers, each of them composed by a message passing round as in Eq. (\ref{message_passing_scheme}), an edge pooling stage as in Eq. (\ref{pooling_scores}) and Eq. (\ref{pooling_scaling}), and a (by-layer) readout function as in Eq. (\ref{local_readout}); (v) a (global) readout function, e.g. as in Eq. (\ref{global_readout}). A schematic view of the whole architecture is illustrated in Figure \ref{fig:tan}. Finally, we present the following result about equivariance properties of the proposed architecture:

\textbf{Theorem 1.} \textit{Cell Attention Networks are permutation invariant.}

In literature, a GNN is permutation invariant if a permutation of  nodes produces the same output without the permutation. More formally, a GNN $f(\cdot)$ taking an input graph $\mathcal{G}$ with adjacency matrix $\mathbf{A}$ and input node features matrix $\mathbf{X} = \{\mathbf{x}_i\}_{i \in \mathcal{V}}$ is (node) permutation invariant if $ f(\mathbf{P}\mathbf{A}\mathbf{P}^T,\mathbf{P}\mathbf{X}) = f(\mathbf{A},\mathbf{X})$ for any permutation matrix $\mathbf{P}$. In the same way, Cell Attention Networks are permutation invariant w.r.t. permutations of nodes, edges and polygons. 

\begin{proof}
We can assert, w.l.o.g., that Attentional Lift is permutation equivariant by construction. The operation $g(\cdot)$ and $a(\cdot)$ are both learnable functions acting on edges, and since $a(\cdot)$ is symmetric by definition, both $a(\cdot)$ and $g(\cdot)$  are symmetric w.r.t. to the vertices that are endpoint of the edges we are considering. This leads to have the whole function permutation equivariant. 

The scheme followed in Edge Pooling is the selecting $top-k$ element of edge set referring to self-attentional coefficients $\gamma_{e}$.
In order to select the $top-k$ elements of $\mathcal{E}$, the vector $\gamma_{e}$ must be sorted. So no matter what is the permutation on the set, after sorting we obtain always the same result. For this reason Edge Pooling is permutation invariant. 

Finally, since the proposed CAN architecture is the composition of a permutation equivariant function (i.e., the attentional lift) and a permutation invariant function (i.e., the edge pooling), it readily follows that CAN are permutation equivariant. Please notice that the proposed architecture without the pooling stage is clearly permutation equivariant.
\end{proof}

\section{Experimental Results}\label{sec:exp_res}

\begin{table}[t]
\begin{center}
\caption{Details of the datasets used in our experiments.}
\label{tab:dataset_details}
\begin{tabular}{lccccc}
\toprule
\multicolumn{1}{c}{Info}  & \multicolumn{1}{c}{MUTAG} & \multicolumn{1}{c}{PTC} & \multicolumn{1}{c}{PROTEINS} & \multicolumn{1}{c}{NCI1} & \multicolumn{1}{c}{NCI109} \\  \bottomrule
\textbf{\# Graphs} & 
\multicolumn{1}{c}{$188$} & 
\multicolumn{1}{c}{$336$} & 
\multicolumn{1}{c}{$1113$} & 
\multicolumn{1}{c}{$4110$} & 
\multicolumn{1}{c}{$4127$} \\
\textbf{\# Classes}            & 
\multicolumn{1}{c}{$2$} & 
\multicolumn{1}{c}{$2$} & 
\multicolumn{1}{c}{$2$}& 
\multicolumn{1}{c}{$2$} & 
\multicolumn{1}{c}{$2$} \\
\textbf{\# Node Feat.}            & 
\multicolumn{1}{c}{$7$} & 
\multicolumn{1}{c}{$20$} & 
\multicolumn{1}{c}{$3$}& 
\multicolumn{1}{c}{$37$} & 
\multicolumn{1}{c}{$38$} \\
\textbf{\# Edge Feat.}            & 
\multicolumn{1}{c}{$4$} & 
\multicolumn{1}{c}{$4$} & 
\multicolumn{1}{c}{$0$}& 
\multicolumn{1}{c}{$0$} & 
\multicolumn{1}{c}{$0$} \\
\textbf{Avg. Nodes}            & 
\multicolumn{1}{c}{$17.93$} & 
\multicolumn{1}{c}{$13.97$} & 
\multicolumn{1}{c}{$39.06$}& 
\multicolumn{1}{c}{$29.87$} & 
\multicolumn{1}{c}{$29.68$} \\
\textbf{Avg. Edges}            & 
\multicolumn{1}{c}{$19.79$} & 
\multicolumn{1}{c}{$14.32$} & 
\multicolumn{1}{c}{$72.82$}& 
\multicolumn{1}{c}{$32.30$} & 
\multicolumn{1}{c}{$32.13$} \\
\textbf{Avg. 3 Cells.}            & 
\multicolumn{1}{c}{$0.00$} & 
\multicolumn{1}{c}{$0.04$} & 
\multicolumn{1}{c}{$27.40$}& 
\multicolumn{1}{c}{$0.04$} & 
\multicolumn{1}{c}{$0.04$} \\
\textbf{Avg. 4 Cells.}            & 
\multicolumn{1}{c}{$0.00$} & 
\multicolumn{1}{c}{$0.01$} & 
\multicolumn{1}{c}{$14.08$}& 
\multicolumn{1}{c}{$0.03$} & 
\multicolumn{1}{c}{$0.03$} \\
\textbf{Avg. 5 Cells.}            & 
\multicolumn{1}{c}{$0.36$} & 
\multicolumn{1}{c}{$0.19$} & 
\multicolumn{1}{c}{$5.68$}& 
\multicolumn{1}{c}{$0.75$} & 
\multicolumn{1}{c}{$0.74$} \\
\textbf{Avg. 6 Cells.}            & 
\multicolumn{1}{c}{$2.5$} & 
\multicolumn{1}{c}{$1.12$} & 
\multicolumn{1}{c}{$8.72$}& 
\multicolumn{1}{c}{$2.66$} & 
\multicolumn{1}{c}{$2.7$} \\
\bottomrule
\end{tabular}
\end{center}
\end{table}

In this section we asses the performance of the proposed architecture when solving several real-world graph classification problems, focusing on well known molecular benchmarks on TUDataset~\cite{TUDataset}. In every experiment, if the dataset is equipped with edge features, we concatenate them to the result of the lift layer (Eq. (\ref{eq:attentional_lift})). We included small molecules with class labels such as \textbf{MUTAG}~\cite{kazius2005derivation} and \textbf{PTC}~\cite{helma2001predictive}. In the former dataset, the task is  to identify mutagenic molecular compounds for potentially commercial drugs, while in the latter the goal is to identify chemical compounds based on their carcinogenicity in rodents. The \textbf{PROTEINS} dataset~\cite{dobson2003distinguishing} is composed mainly by macromolecules. Here, nodes represent secondary structure elements and are annotated by their type. Nodes are connected by an edge if the two nodes are neighbours on the amino acid sequence or one of three nearest neighbors in space; the task is to understand if a protein is an enzyme or not. Using these type of data in a Cell Complex based architecture has an underlying importance since molecules have polyadic structures. Finally, \textbf{NCI1} and \textbf{NCI109} are two datasets aimed at identifying chemical compounds against the activity of non-small lung cancer and ovarian cancer cells~\cite{wale2008comparison}. Considering the aforementioned datasets, we compare CAN with other state of the art techniques in graph representation learning. Since there are no official splits for the training and test sets, to validate the proposed architecture, we followed the method used in~\cite{bodnarcwnet}: we run a 10-fold cross-validation reporting the maximum of the average validation accuracy across folds.

\begin{table}[!ht]
\centering
\caption{Experimental results on TUDatasets. The first part shows the accuracy of graph kernel methods, while the second assess graph neural networks. Cell attention networks scores top on four out of five experiments.}\label{tab:exp-res}
\begin{tabular}{@{}|llllll|@{}}
\toprule
Dataset                         & MUTAG     & PTC      & PROTEINS & NCI1     & NCI109                      \\ \midrule
\multicolumn{1}{|l}{RWK}        & 79.2±2.1  & 55.9±0.3 & 59.6±0.1 & N/A  & \multicolumn{1}{l|}{N/A}      \\
\multicolumn{1}{|l}{GK(k=3)}    & 81.4±1.7  & 55.7±0.5 & 71.4±0.3 & 62.5±0.3 &  \multicolumn{1}{l|}{62.4±0.3} \\
\multicolumn{1}{|l}{PK}         & 76.0±2.7  & 59.5±2.4 & 73.7±0.7 & 82.5±0.5 &  \multicolumn{1}{l|}{N/A}      \\
\multicolumn{1}{|l}{WLK}  & 90.4±5.7  & 59.9±4.3 & 75.0±3.1 & 86.0±1.8 &  \multicolumn{1}{l|}{N/A}      \\ \midrule
\multicolumn{1}{|l}{DCNN}       & N/A       & N/A      & 61.3±1.6 & 56.6±1.0  & \multicolumn{1}{l|}{N/A}      \\
\multicolumn{1}{|l}{DGCNN}      & 85.8±1.8  & 58.6±2.5 & 75.5±0.9 & 74.4±0.5 & \multicolumn{1}{l|}{N/A}      \\
\multicolumn{1}{|l}{IGN}        & 83.9±13.0 & 58.5±6.9 & 76.6±5.5 & 74.3±2.7 &  \multicolumn{1}{l|}{72.8±1.5} \\
\multicolumn{1}{|l}{GIN}        & 89.4±5.6  & 64.6±7.0 & 76.2±2.8 & 82.7±1.7 &  \multicolumn{1}{l|}{N/A}      \\
\multicolumn{1}{|l}{PPGNs}      & 90.6±8.7  & 66.2±6.6 & 77.2±4.7 & 83.2±1.1 &  \multicolumn{1}{l|}{82.2±1.4} \\
\multicolumn{1}{|l}{NGN} & 89.4±1.6  & 66.8±1.7 & 71.7±1.0 & 82.4±1.3 & \multicolumn{1}{l|}{N/A}      \\
\multicolumn{1}{|l}{GSN}        & 92.2±7.5  & 68.2±7.2 & 76.6±5.0 & 83.5±2.0 & \multicolumn{1}{l|}{N/A}      \\
\multicolumn{1}{|l}{SIN}        & N/A       & N/A      & 76.4±3.3 & 82.7±2.1 & \multicolumn{1}{l|}{N/A}      \\
\multicolumn{1}{|l}{CIN}        & 92.7±6.1  & 68.2±7.2 & 77.0±4.3 & 83.6±1.4  & \multicolumn{1}{l|}{$\mathbf{84.0 \pm 1.6}$} \\ \midrule
CAN                      & $\mathbf{94.1}$ ± $\mathbf{4.8}$         & $\mathbf{72.8}$ ± $\mathbf{8.3}$        & $\mathbf{78.2}$ ± $\mathbf{2.0}$       &$\mathbf{84.5}$ ± $\mathbf{1.6}$   &$83.6$ ± $1.2$                     \\ \bottomrule
\end{tabular}
\end{table}


The performance of CAN is reported in Table \ref{tab:exp-res}, along with those of graph kernel methods: Random Walk Kernel (RWK,~\cite{gartner2003graph}), 
Graph Kernel (GK,~\cite{shervashidze2009efficient}),
Propagation Kernels (PK,~\cite{neumann2016propagation}), 
Weisfeiler-Lehman graph kernels (WLK,~\cite{shervashidze2011weisfeiler}); other GNNs: Diffusion-Convolutional Neural Networks (DCNN,~\cite{atwood2016diffusion}),  Deep Graph Convolutional Neural Network (DGCNN,~\cite{zhang2018end}), Invariant and Equivariant Graph Networks (IGN,~\cite{maron2018invariant}), Graph Isomorphism Networks (GIN,~\cite{xu2018powerful}), Provably Powerful Graph Networks (PPGNs,~\cite{maron2019provably}), Natural Graph Networks (NGN,~\cite{de2020natural}), Graph Substructure Network (GSN~\cite{bouritsas2022improving}) and topological networks: Simplicial Isomorphism Network (SIN,~\cite{bodnar2021weisfeiler},
Cell Isomorphism Network (CIN,~\cite{bodnarcwnet}). As we can see from Table \ref{tab:exp-res}, CAN achieves the best performance on four out of five benchmarks, while performing very similarly to CIN in the last experiment (i.e., NCI109). Since CAN has a much lower computational complexity than CIN (cf. Appendix C), these results support the validity and the performance obtained of the proposed architecture. The tested  models have been implemented using PyTorch~\cite{pytorch19}. The datasets have been taken from the PyTorch Geometric library~\cite{pyg2019}. The operations involved during cellular lifting maps use the code provided by \cite{bodnarcwnet} under MIT license.
PyTorch, NumPy, SciPy and are made available under the BSD license, Matplotlib under the PSF license, graph-tool under the GNU LGPL v3 license. PyTorch Geometric is made available under the MIT license. All the experimental results have been made on NVIDIA® GeForce RTX 3090 GPUs  with 10,496 CUDA cores and 24GB GPU memory. The operative system used for the experiment was Ubuntu 22.04 LTS 64-bit. See Appendix \ref{sec:additional_exp} for an extensive description of the tested architectures and an ablation study \footnote{The code implementation for the proposed architecture is available at:  \url{https://github.com/lrnzgiusti/can}}.

\section{Conclusion and Discussion}

In this work we presented \emph{Cell Attention Networks} (CANs),  novel neural architectures operating on data defined over the nodes of a graph incorporated into a
a regular cell complex, exploiting generalized masked self-attention mechanisms. It builds on skeleton-preserving cellular lifting maps, a novel attentional features lift  and a novel edge-level attentional message-passing scheme with two attention functions that operate on the upper and lower connectivities induced by the cell complex.
 The proposed architecture is also equipped with a novel hierarchical edge pooling technique that leverage a self-attention mechanism to downsample the data in the network's hidden layers while extracting significant features for the learning task. The Cell Attention Network architecture proposed and tested in the previous sections shows promising results and it is grounded in the theory of regular cell complexes; however, some directions can be explored to enrich the proposed formulation. In particular, a signal processing perspective~\cite{sardellitti2022cell} can be exploited to reinterpret
 and modify the proposed architecture following a similar approach to~\cite{giusti22}; an expressivity analysis can be carried out based on the renewed Weisfeiler-Lehman approach~\cite{xu2018powerful}, on its generalization to cell complexes~\cite{bodnarcwnet,hajij2022}, or based on spectral approaches~\cite{ribeiro2022aremore}. We leave these problems to be addressed in future works.


\bibliographystyle{unsrtnat}
\bibliography{reference}

\appendix

\section{Experimental Details}\label{sec:additional_exp}

\subsection{Model Implementation}

In our experiments, we employ cell attention networks to regular cell complexes of order two obtained by applying the structural lifting map to the original graphs, i.e. we consider nodes as 0-cells and edges as 1-cells, and the chordless cycles of size up to $R=6$ as 2-cells. In our case, each node of the original graphs is always equipped with an input feature vector.

Throughout all experiments, we employ cell attention networks with the following structure. The attentional lift mechanism in Eq. (\ref{eq:attentional_lift}) is given by:

\begin{align}
    &\mathbf{h}_{e}^0 = \mathbf{x}_e = \overset{F^0}{\underset{k=1}{||}}\phi_n \underbrace{\left( \left(  \left(\mathbf{a}_{n}^{k} \right)^T [\mathbf{x}_i \vert \vert \mathbf{x}_j] \right) \, \big\vert \big\vert \, \tilde{\mathbf{x}}_e \right)}_{a_n^k}, \quad  i,j \in \mathcal{B}(e),
\end{align}

where $\tilde{\mathbf{x}}_e$ is the input feature vector of the edge $e$. If not provided by the specific benchmark, $\tilde{\mathbf{x}}_e$ can be considered as an empty vector. Also, $\mathbf{a}_n^k \in \mathbb{R}^{2F_n}$ is the vector of attention coefficients associated to the k-th feature of the input edge feature vector, and $\phi_n$ is the non-linear activation function for the lift layer. Please notice that the employed functions $a_n^k$ are not symmetric, but they give the best learning performance on the proposed tasks.

The lower and upper attentional  functions $a_d(\mathbf{h}_{e},\mathbf{h}_{k})$ and $a_u(\mathbf{h}_{e},\mathbf{h}_{k})$ in Eq. (\ref{message_passing_scheme}) are chosen as two independent masked self-attention schemes. They can be chosen following any of the known approaches from graphs~\cite{velivckovic2017graph, brody2021attentive}. In this paper we follow the approach from \cite{velivckovic2017graph}: formally, let

\begin{align}
    &\omega_{e,k}^{l,d} =  \phi_a \left( (\mathbf{a}_{d}^l)^T \left[ \mathbf{W}_{d}^{l}  \mathbf{h}_{e}^{l}  \vert \vert   \mathbf{W}_{d}^{l} \mathbf{h}_{k}^{l}  \right] \right)\label{eq:low_omega} \\
    &\omega_{e,k}^{l,u} = \phi_a \left( (\mathbf{a}_{u}^l)^T \left[ \mathbf{W}_{u}^{l}  \mathbf{h}_{e}^{l}  \vert \vert   \mathbf{W}_{u}^{l} \mathbf{h}_{k}^{l}  \right] \right), \label{eq:up_omega}
\end{align}

where $\mathbf{a}_{d}^l, \mathbf{a}_{u}^l \in \mathbb{R}^{F^{l}}$ are two independent vectors of attention coefficients, $\mathbf{W}_{d}^l, \mathbf{W}_{u}^l \in \mathbb{R}^{F^{l+1} \times F^{l}}$ are two learnable linear transformations shared by the lower and upper neighbourhoods of the complex, respectively, and $\phi_a$ is a pointwise non-linear activation. The coefficients $\omega^u$ and $\omega^d$ in Eq. (\ref{eq:low_omega},\ref{eq:up_omega}) represent the importance of the features of edge k when exchanging messages with edge e over lower and upper neighborhoods, respectively. It worth to emphasize that since the attention schemes are decoupled, these importance coefficients will be different over the upper and lower neighborhoods. 

In line with the approach of \cite{velivckovic2017graph}, we make coefficients easily comparable across different edge by normalizing them across all choices of $k$ using the softmax function:

\begin{align}
    &\alpha_{e,k}^{l,d} = \frac{\exp\left( \omega_{e,k}^{l,d} \right) } 
    {\sum_{\iota \in \mathcal{N}_{d}^{l}(e)} \exp\left( \omega_{e,\iota}^{l,d}   \right)} \\[5pt]
    &\alpha_{e,k}^{l,u} = \frac{\exp\left( \omega_{e,k}^{l,u} \right) } 
    {\sum_{\iota \in \mathcal{N}_{u}^{l}(e)} \exp\left( \omega_{e,\iota}^{l,u}   \right)}
\end{align}

Thus, for layer $l$ we have that $a^l_d(\mathbf{h}_{e}^l,\mathbf{h}_{k}^l) = \alpha_{e,k}^{l,d}$ and $a^l_u(\mathbf{h}_{e}^l,\mathbf{h}_{k}^l) = \alpha_{e,k}^{l,u}$. Once the attention coefficients have been normalized, to update the representation of an edge $e$, a linear combination of the edge features and the normalized attention coefficients corresponding to them is computed for both the lower and upper neighbourhoods and the results are aggregated alongside with the current edge representation. Formally:

\begin{align}
    &\widetilde{\mathbf{h}}_{e}^{l} = \phi\Bigg((1+\varepsilon) \, \mathbf{W}_{s}^{l} \mathbf{h}_{e}^{l} + \sum_{k \in \mathcal{N}^l_d(e)} \alpha_{e,k}^{l,d} \, \mathbf{W}_{d}^{l} \mathbf{h}_{k}^l + \sum_{k \in \mathcal{N}^l_u(e)} \alpha_{e,k}^{l,u} \, \mathbf{W}_{u}^{l} \mathbf{h}_{k}^l)\Bigg),
\end{align}

here $\mathbf{W}_{s}^l \in \mathbb{R}^{F^{l+1} \times F^{l}}$ is a shared linear transformation applied to the current hidden representation of the edges of the complex. Notice that the functions $\psi_d(\mathbf{h}_{k}^l)$ and $\psi_u(\mathbf{h}_{k}^l)$ in Eq. (\ref{message_passing_scheme}) are implemented respectively as: $\mathbf{W}_{d}^{l} \mathbf{h}_{k}^l$ and $\mathbf{W}_{u}^{l} \mathbf{h}_{k}^l$.

In the pooling layer, the hidden feature vectors are updated using Eq. (\ref{pooling_scaling}) by scaling the features $\widetilde{\mathbf{h}}_{e}^{l}$ with the corresponding score $\gamma_e^l$:

\begin{align}
    &\mathbf{h}_{e}^{l+1} = \underbrace{\phi_p \left( ( \mathbf{a}_p^l )^T \tilde{\mathbf{h}}_e^l  \right)}_{\gamma_e^l}  \tilde{\mathbf{h}}_e^l, \quad \forall e \in \mathcal{E}^{l+1},
\end{align}

where the vector $\mathbf{a}_p^l$ plays the role of a collection of attention coefficient that weight the features of $\tilde{\mathbf{h}}_e^l$  to compute the corresponding score $\gamma_e^l$, which that represents the importance of edge $e$ in the learning task. Following the approach of \cite{lee2019self}, the weight  $( \mathbf{a}_p^l )^T \tilde{\mathbf{h}}_e^l$ of the edge $e$ is also forwarded into a non-linear activation function $\phi_p$ to produce the score $\gamma_e^l \in \mathbb{R}$, which is then multiplied to $\tilde{\mathbf{h}}_e^l$ to obtain $\mathbf{h}_{e}^{l+1}$.

Readout operations are performed as follows: If the pooling approach is hierarchical, the readout is performed layer-wise. In particular, we choose the sum as the \emph{permutation equivariant aggregation function} of Eq. (\ref{local_readout} which results in a \emph{hierarchical representation} of the complex i.e. a collection $\{\mathbf{h}_{\mathcal{C}^l}\}_{l=0}^{L-1}$ of hidden representations. Then, Eq. (\ref{global_readout}) is computed as the sum over the collection defined previously.

In the case of a global pooling, the readout is computed only in the last layer, which constitute  the overall representation of the complex:

\begin{align}
    \mathbf{h}_{\mathcal{C}}= \mathbf{h}_{\mathcal{C}^{L-1}} = \sum_{e \in \mathcal{E}^{L-1}} \mathbf{h}_{e}^{L-1}.
\end{align}

Once $\mathbf{h}_{\mathcal{C}}$ is obtained, it is forwarded into a 2-Layer MLP with $\phi$ as activation function to perform the prediction.

In all layers, we adopt a Batch Normalization technique~\cite{ioffe2015batch} and all training operations are performed with the AdamW optimization algorithm~\cite{loshchilov2017decoupled}. In Table \ref{tab:hyper-params} we report the hyper-parameters used in our experiments for each dataset.

\begin{table}[t]
\centering
\caption{Hyperparameter used for the experiments on TUDatasets.}
\label{tab:hyper-params}
\begin{tabular}{lccccc}
\toprule 
\multicolumn{1}{l}{Parameter}  & \multicolumn{1}{c}{MUTAG} & \multicolumn{1}{c}{PTC} & \multicolumn{1}{c}{PROTEINS} & \multicolumn{1}{c}{NCI1} & \multicolumn{1}{c}{NCI109} \\ \bottomrule
Lift Heads                   & \multicolumn{1}{c}{$1$} & \multicolumn{1}{c}{$32$} & \multicolumn{1}{c}{$256$} & \multicolumn{1}{c}{$107$} & \multicolumn{1}{c}{$116$} \\
Lift Activation            & \multicolumn{1}{c}{\textit{ReLU}} & \multicolumn{1}{c}{\textit{ELU}} & \multicolumn{1}{c}{\textit{ELU}} & \multicolumn{1}{c}{\textit{ELU}} & \multicolumn{1}{c}{\textit{Sigmoid}} \\
Lift Dropout                    & \multicolumn{1}{c}{$0.0$} & \multicolumn{1}{c}{$0.0$} & \multicolumn{1}{c}{$0.05$} & \multicolumn{1}{c}{$0.2$} & \multicolumn{1}{c}{$0.2$}\\
Hidden Features           & \multicolumn{1}{c}{$[32,32]$} & \multicolumn{1}{c}{$[32,8]$} & \multicolumn{1}{c}{$[128,128]$} & \multicolumn{1}{c}{$[32,16,64,8]$} & \multicolumn{1}{c}{$[64,8,8,32,8]$} \\
Attention Heads             & \multicolumn{1}{c}{$[1,1]$} & \multicolumn{1}{c}{$[2,1]$} & \multicolumn{1}{c}{$[1,1]$} & \multicolumn{1}{c}{$[3,5,4,5]$} & \multicolumn{1}{c}{$[5,7,4,7,7]$} \\
Attention Aggregation      & \multicolumn{1}{c}{\textit{-}} & \multicolumn{1}{c}{\textit{cat}} & \multicolumn{1}{c}{\textit{-}} & \multicolumn{1}{c}{\textit{cat}} & \multicolumn{1}{c}{\textit{cat}} \\
Attention Activation       & \multicolumn{1}{c}{\textit{LReLU}} & \multicolumn{1}{c}{\textit{LReLU}} & \multicolumn{1}{c}{\textit{Tanh}} & \multicolumn{1}{c}{\textit{Tanh}} & \multicolumn{1}{c}{\textit{Tanh}} \\
Activation                & \multicolumn{1}{c}{ELU} & \multicolumn{1}{c}{ELU} & \multicolumn{1}{c}{Tanh} & \multicolumn{1}{c}{ELU} & \multicolumn{1}{c}{GELU} \\
MLP Neurons                & \multicolumn{1}{c}{$8$} & \multicolumn{1}{c}{$4$} & \multicolumn{1}{c}{$128$} & \multicolumn{1}{c}{$256$} & \multicolumn{1}{c}{$32$} \\
Batch Size                 & \multicolumn{1}{c}{$64$} & \multicolumn{1}{c}{$128$} & \multicolumn{1}{c}{$128$} & \multicolumn{1}{c}{$128$} & \multicolumn{1}{c}{$128$}\\
Neg. Slope                & \multicolumn{1}{c}{$0.1$} & \multicolumn{1}{c}{$0.1$} & \multicolumn{1}{c}{$0.3$} & \multicolumn{1}{c}{$0.08$} & \multicolumn{1}{c}{$0.07$}\\
Pool Ratio         & 
\multicolumn{1}{c}{$1.0$} & \multicolumn{1}{c}{$0.75$} & \multicolumn{1}{c}{$0.6$} & \multicolumn{1}{c}{$0.5$} & \multicolumn{1}{c}{$0.75$} \\
Pool Type               & \multicolumn{1}{c}{\textit{Hier.}} & \multicolumn{1}{c}{\textit{Glob.}} & \multicolumn{1}{c}{\textit{Hier.}} & \multicolumn{1}{c}{\textit{Glob.}} & \multicolumn{1}{c}{\textit{Glob.}} \\
Dropout                    & \multicolumn{1}{c}{$0.1$} & \multicolumn{1}{c}{$0.6$} & \multicolumn{1}{c}{$0.3$} & \multicolumn{1}{c}{$0.15$} & \multicolumn{1}{c}{$0.05$}\\
Learning Rate               & \multicolumn{1}{c}{$3e-3$} & \multicolumn{1}{c}{$1e-3$} & \multicolumn{1}{c}{$3e-3$} & \multicolumn{1}{c}{$3e-4$} & \multicolumn{1}{c}{$3e-3$}    \\
\bottomrule
\end{tabular}
\end{table}

\subsection{Complexity Analysis}
In this section, we review the computational complexity of each operation involved the proposed architecture referring to the model implementation define above:

\textbf{Cellular Lifting Map}: Although this operation can be precomputed for the entire dataset and the connectivity results stored for a later usage, it worth to elicit its complexity noticing that for some applications the storage of the upper and lower connectivity for the \emph{entire} dataset might be not possible. We considered Cellular Lifting Maps that assign 2-cells to all the chordless cycles of a graph with a maximum number of nodes in the cycles up to $R$ as maximum cycle size. The chord-less cycles in a graph can thus be enumerated in $\mathcal{O}((|E| + |V| R) \, \textrm{polylog} |V|)$ time~\cite{ferreira14}. Similar to~\cite{bodnarcwnet}, in our experimental setup we have that $R$ can upper bounded by a small constant. Thus, the complexity of this operation can be approximated to be linear in the size of the complex.

\textbf{Attentional Lift}: The complexity of this operation consists of a multi-head attention message passing scheme over the entire graph~\cite{velivckovic2017graph}. For a single node pair $i,j \in \mathcal{V}$ connected by an edge $e \in \mathcal{E}$, the attentional lift defined in Eq. (\ref{eq:attentional_lift}) can be decomposed into $F^0$ independent self-attention schemes. Each attention scheme requires $\mathcal{O}(F_n)$ computations, where $F_n$ is the number of input node features. Thus, for the pair $i,j$, the attentional lift is performed in $\mathcal{O}(F^0 F_n)$, where $F^0$ is a parameter to be chosen as the number of input edge features. Accounting all the edges of the complex yields an amount of $\mathcal{O}(\vert \mathcal{E} \vert F^0 F_n))$ operations to lift the given node features into edge ones.

\textbf{Cell Attention Layer}: This operation consists in two independent masked self-attention message passing schemes over the upper and lower neighbourhoods of the complex, namely cell attention, an inner linear transformation of the edges' features and an outer point-wise nonlinear activation (Eq.(\ref{message_passing_scheme})). For the lower neighbourhood, a single edge $e$ receives at most $2$ messages for each cell in its coboundary, $\mathcal{CB}(e)$ (cf Fig. \ref{fig:att}.1 ); thus, for a single edge the attention over the lower neighbourhood of the complex is $\mathcal{O}(\vert \mathcal{CB}(e) \vert)$, where $\vert \mathcal{CB}(e) \vert$ is the number of cells that are co-faces of the edge $e$. Regarding the upper neighbourhood, if $R$ is the maximum ring size we have that a single edge $e$ receives $\mathcal{O}(R \cdot \vert \mathcal{CB}(e) \vert)$ messages (cf. Fig. \ref{fig:att}.2. Recalling that $R$ is upper bounded by a small constant~\cite{bodnarcwnet}, cell attention is an $\mathcal{O}(\vert \mathcal{CB}(e) \vert)$ operation for both neighbourhoods of an arbitrary edge $e$ i.e. linear in the size of the complex. The inner linear transformation that propagates the information contained in $\mathbf{h}^{l}_e$ is upper bounded by $\mathcal{O}(F^{l})$. Extending this to all edges of the complex, we have that the complexity of a cell attention layer can be rewritten as $\mathcal{O}(|\mathcal{E}^l|\, F^{l})$. In the case of a multi-head cell attention, the complexity receives an overhead induced by the number of attention heads involved within the layer, i.e., a multiplication by a factor $H_c$, the number of cell attention heads.

\textbf{Attentional Pooling}: The operations involved in the pooling layer can be decomposed in: (i) computing the self-attention scores for each edge of the complex ($\gamma_e^l$ in  Eq. (\ref{pooling_scores})); (ii) select the highest $\lceil k \, \vert \mathcal{E}^{l} \vert \, \rceil$ values from a collection of self-attention scores ($\text{top-k}(\{\gamma^l_e\}_{e \in \mathcal{E}^l}, \lceil k |\mathcal{E}^l| \rceil)$); and (iii) adjust the connectivity of the complex (see Fig. \ref{fig:pooling}). To compute the computational complexity of this layer it is convenient to see the selection operation as a combination of a sorting algorithm over a collection of self-attention scores and a selection of the first $\lceil k \, \vert \mathcal{E}^{l} \vert \, \rceil$ elements from the sorted collection. Since the computations involved in (i) and (iii) are linear in the dimension of the complex, the overall complexity of this layer in can be upper bounded by the sorting algorithm, i.e., $\mathcal{O}(|\mathcal{E}^{l}| \; log (|\mathcal{E}^{l}|))$.

In practice, all the computations involved in a cell attention network are local formulations completely disjoint from each other. Thus, using an efficient GPU-based implementation, we can rewrite \emph{all} the analysis in terms of the longest sequential chain of operations in a concurrent execution over the edges of the input domain, i.e., $\mathcal{O}(log(|E|))$. For an in-depth concurrency analysis we refer readers to~\cite{Besta22parallel}, where the authors report a complete taxonomy of parallelism in GNNs.   

\subsection{Learnable Parameters}

The total number of learnable parameters of a CAN can be decomposed into:

\textbf{Cellular Lifting Map}: Lifting the input graph $\mathcal{G}$ to a cell complex $\mathcal{C}$ is an operation that assign a cell $\sigma$ to \emph{all} the chord-less cycles of $\mathcal{G}$ up to a maximum cycle size $R$. Intuitively this operation does not involve any parameter to be learned during the network's training phase. Thus, the number of learnable parameters for the lift is $\mathcal{O}(1)$.

\textbf{Attentional Lift}: In the context of lifting a pair of graph node features $\mathbf{x}_i, \mathbf{x}_j \in \mathbb{R}^{F_n}$ to a signal defined over the edge of the complex, $\mathbf{x}_{e} \in \mathbb{R}^{F^0}$, we have to learn a vector of attention coefficients $\mathbf{a}_{n} \in \mathbb{R}^{2 F_n}$ for each input edge feature. The vector $\mathbf{a}_{n}$ has a number of learnable parameters in $\Theta \left( F_n \right)$. Accounting multiple input edge features, the overall number of parameters for the attention lift operation is $\Theta(F^0 F_n)$, where $F^0$ is the number of input edge features computed as multiple independent attention heads.

\textbf{Cell Attention}: In terms of learnable parameters, a single \emph{cell attention layer} is composed of: two independent vectors of attention coefficients $\mathbf{a}_d^l, \mathbf{a}_u^l \in \mathbb{R}^{F^{l}}$ for properly weighting the lower and upper neighbourhoods, respectively. Moreover, the layer is equipped with three linear transformations, $\mathbf{W}_s^l, \mathbf{W}_d^l, \mathbf{W}_u^l \in \mathbb{R}^{F^{l} \times F^{l+1}}$ acting respectively on: $\mathbf{h}_e^l$, the hidden feature vector of edge $e$ at layer $l$ and the hidden feature vectors $\mathbf{h}_k^l$ in the lower and upper neighbourhoods of the edge $e$. Thus, the number of learnable parameters of a cell attention layer is $\mathcal{O}(F^{l} F^{l+1})$.

\textbf{Pooling}: For this layer, learnable parameters are employed only in computing the self-attention scores ($\gamma_e^l$, Eq. (\ref{pooling_scores})). The shared vector of attentional scores' coefficients $\mathbf{a}_p^l \in \mathbb{R}^{F^{l}}$, similarly to the lift layer, is known to have a number of learnable parameters in $\Theta \left( F^{l} \right)$.

\subsection{Ablation Study}\label{sec:ablation}

In this section we take a detailed look at the performance of each operation involved in cell attention networks by performing different ablation studies and show their individual importance and contributions.

\begin{figure}[!htb]
    \centering
    \includegraphics[width=.8\textwidth]{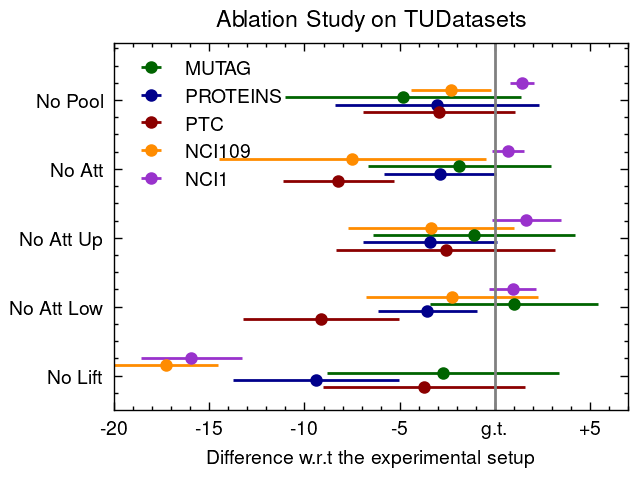}
    \caption{TUDataset: Results of the ablation of different CAN features with respect to Table \ref{tab:exp-res} (g.t.). The ablation study shows the benefits of incorporating all the proposed operations into the message passing procedure when operating on data defined over cell complexes.}
    \label{fig:ablation}
\end{figure}

In particular, we followed the same experimental setup used in section \ref{sec:exp_res} by fixing the hyper-parameters as in Table \ref{tab:hyper-params} and removing one-by-one the cell attention network operations: (i) removing the lift refers to assign a feature $\mathbf{x}_e$ to an edge $e$ using a linear function that takes the feature vectors $\mathbf{x}_i, \mathbf{x}_j$  from the vertices $i,j \in \mathcal{B}(e)$, i.e. a simple scalar product between $\mathbf{x}_i$ and $\mathbf{x}_j$ ($\mathbf{x}_e = \langle \mathbf{x}_i, \mathbf{x}_j \rangle$); (ii) removing the lower attention can be intended as initializing the lower attention coefficients as: $\mathbf{a}_{d}^l = \boldsymbol{1}$ and left them unmodified during the update step in the training phase; (iii) similarly, for the upper attention we replicated the idea of (ii) but now only the upper attention coefficients are involved, i.e.  $\mathbf{a}_{u}^l = \boldsymbol{1}$ and kept fixed for the entire optimization stage; (iv) removing the attention means to remove both the upper and lower attention simultaneously as explained in (ii) and (iii); removing the pooling means to detach the pooling layer from the network and remove eventual intermediate readout computations involved in the hierarchical pooling setup.

As shown in Figure \ref{fig:ablation} and in Table \ref{tab:ablation}, we observe a decrease in the overall performance when removing parts of the cell attention network architecture as expected. Of particular interest is the ablation study on NCI1, which shows a slightly higher accuracy in every case we kept the attention coefficients fixed and without the pooling but a drastic drop in the performance when the edge features are no longer learned. Moreover we see that there are no evident "patters" inside the ablation study with the except that for NCI109 we observe the same behaviour of NCI1 when removing the lift layer. This fact can be explained by noticing that the aforementioned datasets experience, on average, a very similar topology (Table \ref{tab:dataset_details}).

\begin{table}[t]
\centering
\caption{Analysis of the impact of the operations involved in cell attention networks.}
\label{tab:ablation}
\begin{tabular}{lccccc}
\toprule
\multicolumn{1}{l}{Feature Removed}  & \multicolumn{1}{c}{MUTAG} & \multicolumn{1}{c}{PTC} & \multicolumn{1}{c}{PROTEINS} & \multicolumn{1}{c}{NCI1} & \multicolumn{1}{c}{NCI109} \\
\bottomrule 
No Pooling                   & \multicolumn{1}{c}{$-4.8 \pm 6.2 $} & \multicolumn{1}{c}{$-2.9 \pm 3.9$} & \multicolumn{1}{c}{$-3.0 \pm 5.3$} & \multicolumn{1}{c}{$+1.4 \pm 0.6$} & \multicolumn{1}{c}{$-2.3 \pm 2.1$} \\
No Attention            & \multicolumn{1}{c}{$-1.9 \pm 4.8$} & \multicolumn{1}{c}{$-8.2 \pm 2.9$} & \multicolumn{1}{c}{$-2.9 \pm 2.9$ } & \multicolumn{1}{c}{$+0.7 \pm 0.8$} & \multicolumn{1}{c}{$-7.5 \pm 7.0$} \\
No Upper Attention                    & \multicolumn{1}{c}{$-1.1 \pm 5.3$} & \multicolumn{1}{c}{$-2.5 \pm 5.7$} & \multicolumn{1}{c}{$-3.4 \pm 3.5$} & \multicolumn{1}{c}{$+1.6 \pm 1.8$} & \multicolumn{1}{c}{$-3.3 \pm 4.3$}\\
No Lower Attention           & \multicolumn{1}{c}{$+1.0 \pm 4.4$} & \multicolumn{1}{c}{$-9.1 \pm 4.1$} & \multicolumn{1}{c}{$-3.5 \pm 2.6$} & \multicolumn{1}{c}{$+0.9 \pm 1.2$} & \multicolumn{1}{c}{$-2.2 \pm 4.5$} \\
No Lift             & \multicolumn{1}{c}{$-2.7 \pm 6.1$} & \multicolumn{1}{c}{$-3.7 \pm 5.2$} & \multicolumn{1}{c}{$-9.4 \pm 4.3$} & \multicolumn{1}{c}{$-15.9 \pm 2.6$
} & \multicolumn{1}{c}{$-17.3 \pm 2.7$} \\
\bottomrule
\end{tabular}
\end{table}

\end{document}